# Detection and Recognition of Malaysian Special License Plate Based on SIFT Features

Hooi Sin Ng, Yong Haur Tay
Faculty of Engineering Science,
Universiti Tunku Abdul Rahman (UTAR),
Jalan Genting, Kelang,
53300, Setapak,
Kuala Lumpur
email: ng_hooisin@yahoo.com, tayyh@utar.edu.my

Kim Meng Liang, Hamam Mokayed, Hock Woon Hon
Advanced Informatics Architecture Development (AIAD)
MIMOS Berhad,
Technology Park Malaysia,
57000, Kuala Lumpur
email: liang.kimmeng@mimos.my, hamam.mokayed@mimos.my, hockwoon.hon@mimos.my

Abstract – Automated car license plate recognition systems are developed and applied for purpose of facilitating the surveillance, law enforcement, access control and intelligent transportation monitoring with least human intervention. In this paper, an algorithm based on SIFT feature points clustering and matching is proposed to address the issue of recognizing Malaysian special plates. These special plates do not follow the normal standard car plates' format as they may contain italic, cursive, connected and small letters. The algorithm is tested with 150 Malaysian special plate images under different environment and the promising experimental results demonstrate that the proposed algorithm is relatively robust.

## I. Introduction

Automatic License Plate Recognition (ALPR) is one of the most important applications of Intelligent Transportation System (ITS). It can be extensively applied to stolen license plate tracking, roadside inspection, parking lot administration, or automotive factory management with least human intervention.

In many developed countries, the license plates are issued by the government department of road transport whereby the plate number are standardized in terms of plate's size, the ordering, the font and the spacing between characters. In Malaysia, the standard format of license plate for private and commercial vehicles consists of two parts where it started with alphabets and ended by numbers. However, there are some exceptions to the standard format plate. One of the exceptions is the special license plates (as shown in Figure 1) which have been introduced from time to time to commemorate certain events or occasions.

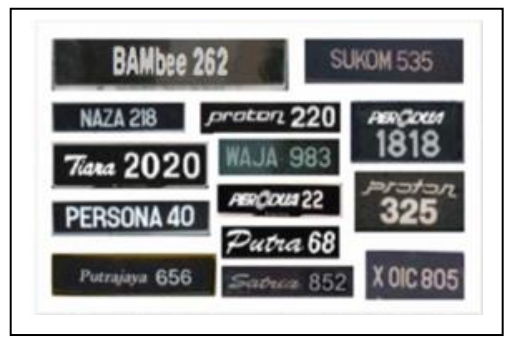

Figure 1: Samples of Special Malaysian License Plate

These kinds of plates bearing distinctive prefixes were made available by the Malaysian Road Transport Department at a higher cost. For example, when a new local car model is launched, the plates of first one thousand owners will bear the prefix of the car model such as Proton, Perodua, Satria, Bambee, Persona, and Waja. The ALPR systems available in the market are not able to recognize these kinds of plates due to their special properties. These plates normally bearing prefixes where the letters are written in small case, cursive, italic, and highly connected. These prefixes need to be handled separately for recognition as they are difficult to be segmented (as shown in Figure 2) into separate characters like other standard plates.

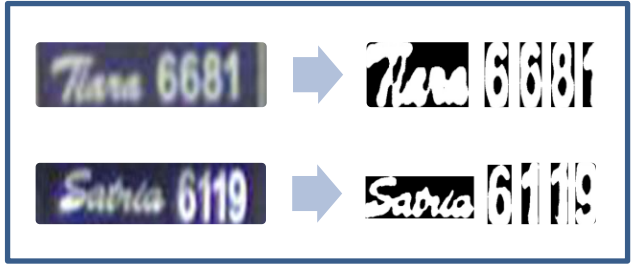

Figure 2: Segmentation of Special Malaysian License Plate

This paper is organized as follows. Section 2 presents the related works. Section 3 describes special plates recognition algorithm based on SIFT feature. Section 4 presents the experimental results and conclusion is given in Section 5.

## II. Related Works

A lot of researches have been done on numbers and letters recognition. Those recognition methods can be generalized into two main categories, which is template based method and supervised learning-based method.

Template based method [8-11] is the typical method used in character recognition. The method compares the segmented characters with stored templates. To recognize the character, the input character is compared to each template to find either an exact match or the closest representation between template and input character. Normalized correlation method is the common technique used for matching process whereby a value indicate how well a template matches with the input character will be returned. However, this method is sensitive to image orientation and noise disturbance. There are several mature algorithms have been proposed and widely used like supervised learning based method.

For supervised learning-based method, two commonly used classifiers for character recognition are artificial neural network (ANN) [15-17] and support vector machine (SVM) [12-14]. ANN is one of the powerful classifiers. It is essentially a mapping relationship where it expresses the nonlinear mapping relation between the input unit and output unit. Performance of ANN depends on size, type of training samples and feature extraction, but SVM can run effectively and efficiently with small samples and used pixel values as vectors directly [2].

Scale invariant feature transform (SIFT) descriptor, a highly distinctive invariant, is proposed by D. Lowe in 2004 [18]. SIFT descriptor is proved to be effective and robust in a lots of pattern recognition applications where local features are critical and vital for the algorithms correct response [1]. Other general applications include object recognition, robotic mapping and navigation, image stitching, 3D modeling, gesture recognition, video tracking.

Recently, SIFT algorithm has been applied in Chinese character recognition [2, 4, 7], Farsi/Arabic font recognition [3] and license plate detection [1, 5]. An excellent recognition rate of nearly 100% is obtained in [3] using a database of 1400 text images. In [2], SIFT feature points clustering and matching in which a center matching strategy is proposed to recognize Chinese character for LPR application. In [5], SVM trained by SIFT feature to detect letter and number in order to reject noise candidate has reached recall rate and precision rates of 92% and 90% respectively.

Due to the main characteristics of the SIFT features like robustness against changes of illumination, scaling, rotation and affine distortion [1] and high accuracy result in previous works, the SIFT technique is expected to give good recognition rate for recognition of Malaysian special license plate.

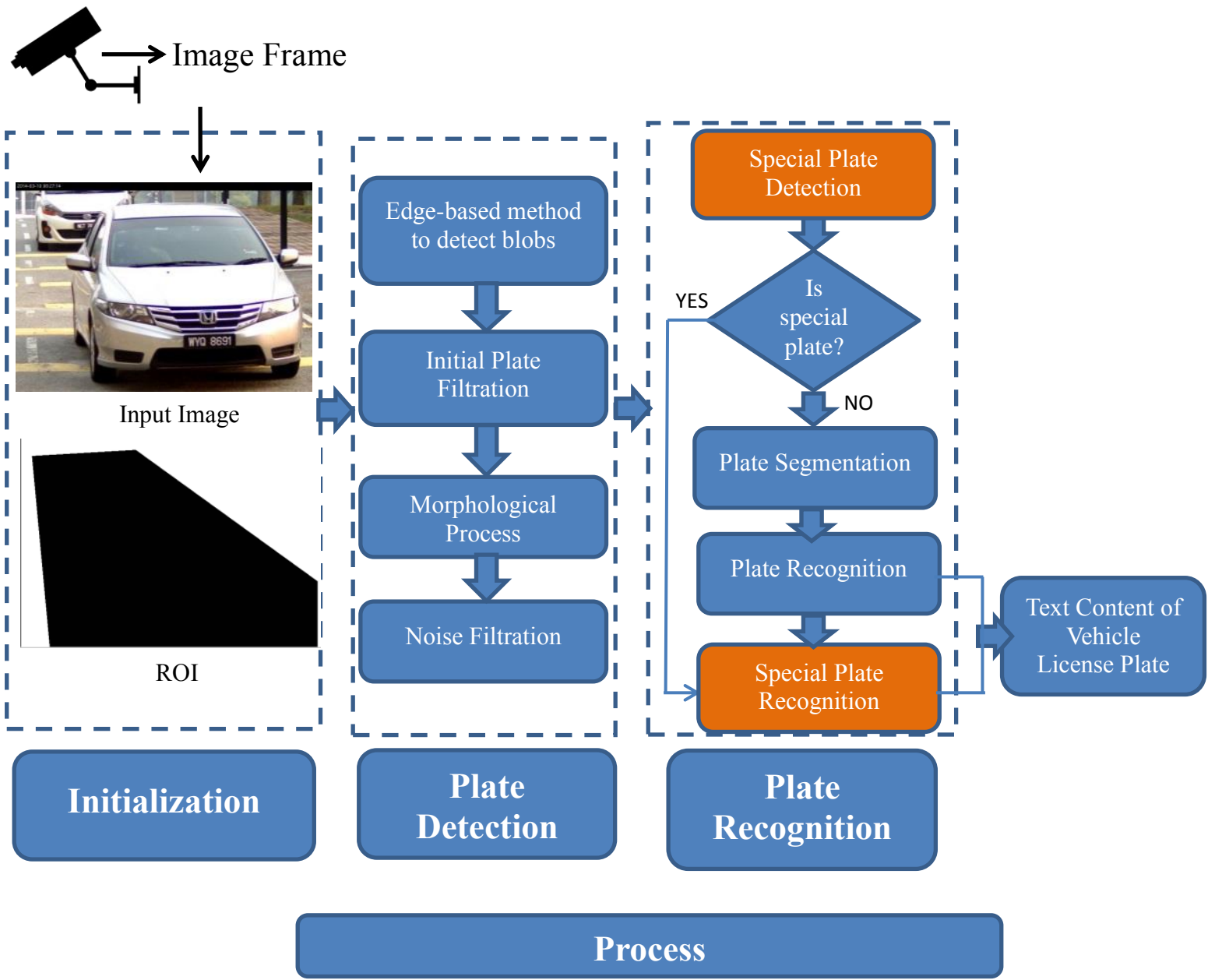

Figure 3: Overview of Overall LPR System

To illustrate the performance of proposed framework, frame image which includes a car is illustrated in Figure 3 along with the overall LPR system. The LPR system consists of three main processes, namely initialization stage, plate detection stage and plate recognition stage. In the initializations process, the objects that do not exist in the region of interest (ROI) will be removed.

In the plate detection stage, a few of operations are implemented. Firstly, edge detections with Sobel operators is performed to detect the vertical edges and thresholding is then applied on the output of the sobel operator. Secondly, morphology processing is applied to dilate the vertical edges horizontally into a blob. Next, filtering on the blobs is performed to minimize false plate to be detected by using area of the blob, height and width ratio, compactness and ratio white pixel/black pixel of the binarized blob sub-image, then those remaining blobs after filtering process are considered as potential license plate to be passed into recognition stage.

In the plate recognition stage, special plate recognition process is implemented to detect and recognize whether the input license plate image is special plate, which will be discussed in this paper in the following section. If the input plate image is normal plate, plate segmentation process is performed to separate each of the characters into one single blob. Each of these blobs will then be recognized using neural network classifier and give the final output.

### III. Proposed Methodology

The proposed algorithm for recognition of Malaysian special plates consists of four steps, namely (1) image preprocessing, (2) SIFT feature extraction, (3) keypoints matching and (4) object localization as shown in Figure 4. The location and group of prefix of special license plate is given as the end result.

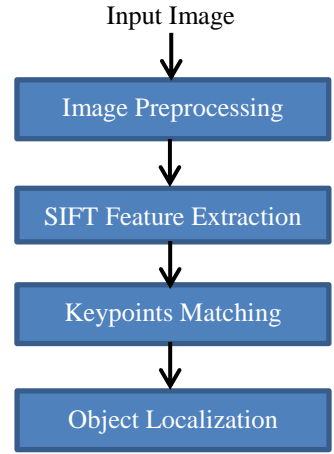

Figure 4: Process Flow of Proposed Algorithm to Detect and Recognize Special Malaysian License Plate

### 3.1 Image Preprocessing

In the image preprocessing stage, the detected license plate image is converted into grayscale image. Soft edges of image are then removed to improve the quality of matching points.

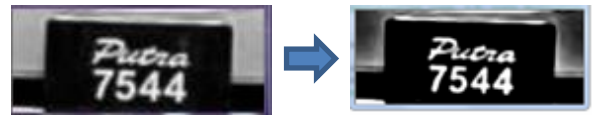

Figure 5: Image before and after preprocessing stage

### 3.2 SIFT Feature Extraction

The next stage is to detect SIFT keypoints of input plate image. SIFT algorithm finds and describes local features in images to help matching the different views of a certain object in the given image. Feature description is formed by interesting points of the object called keypoints. These features are invariant to changes in scaling and rotation, and robust to illumination variation, noise and affine distortion.

The SIFT feature extraction algorithm consists of several filtering stages. The first stage which named as scale-space extrema detection filtered out sets of locations and scales recognizable in different views of the same object with a Gaussian scale-space function. Then difference of Gaussian (DoG) function is calculated by computing the difference between two images. DoG is then applied on a set of re-sampled images to obtain the extreme points. To improve confidence in the final keypoints, low contrast points and poorly localized points on the edges are disposed. The best match for each keypoint in the input image is found by finding its nearest neighbor in the template image by computing the Euclidean distance from the given descriptor vector. A least square-based method is used to verify each identified group of features and finally to discard the outliers of remaining points. If there are less than remaining points after verification stages, the match will be rejected.

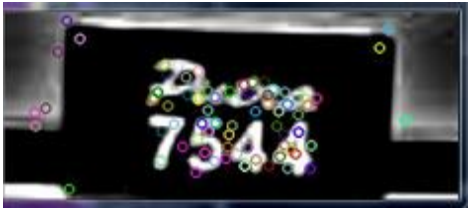

Figure 5: Input image with detected SIFT keypoints

### 3.3 Keypoints Matching

SIFT feature points are stored according to each special prefix in template. Recognition process is performed by comparing a set of extracted feature points between the detected plate and each of the templates based on their Euclidean distance as a common measure of their vectors. After matching two images one gets a vector of good matches which are subsets of keypoints that agree on the template and its location, scale and orientation in the input image. Along with that, two sets of keypoints obtained from template and input image will be used in the matching process.

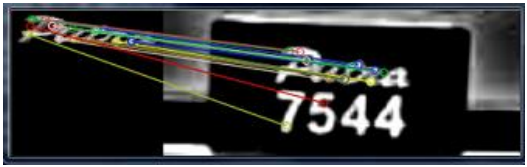

Figure 5: Matched keypoints between template and input image

### 3.4 Object Localization by Estimating Homography

To localize the object, homography between matched keypoints that relates two images is estimated. Homography represents a 3x3 matrix through mapping relationship between the points of the reference and the target images as:

$$X = HX'  \quad (1)$$

Where X and X' represent the reference points and their corresponding target points respectively.

$$H = \begin{bmatrix} h_{11} & h_{12} & h_{13} \\ h_{21} & h_{22} & h_{23} \\ h_{31} & h_{32} & h_{33} \end{bmatrix} \quad (2)$$

A homography is estimated using the DLT (Direct Linear Transform) algorithm between reference image and target image. It can be written as:

$$\begin{bmatrix} 0^T & w_i X_i^T & y_i X_i^T \\ w_i X_i^T & 0^T & x_i X_i^T \end{bmatrix} \begin{bmatrix} h^1 \\ h^2 \\ h^3 \end{bmatrix} = 0 \quad (3)$$

Where $h^j$ represents the j-th column of H, and $X_i^T$ denotes a point $(x_i, y_i, w_i)$ of the reference images. While a point ($x_i^t\ y_i^t\ w_i^t$) represents the corresponding point of the target image.

The highlighted green bounding box shown in Figure 5 below is the matched object in the license plate image.

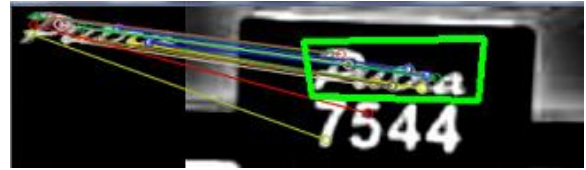

Figure 5: Result of Matched Template in License Plate after Homography Estimation Process

By using the proposed method, the location of the special prefix on the plate is known. The numeric region of the plate is then recognized using the conventional technique like neural network to give the final output to the system.

## IV. Experimental Results

The algorithm has been tested with 500 license plate images which consist of 150 special license plate images and 350 standard plate images captured during daylight at four different locations. There are six types of special license plate being tested in this paper, which are *Perodua, Proton, Satria, Tiara, Putrajaya* and *Putra*. This is because these kinds of special plates bearing prefixes that are Figure 6 shows some sample images used during testing process.

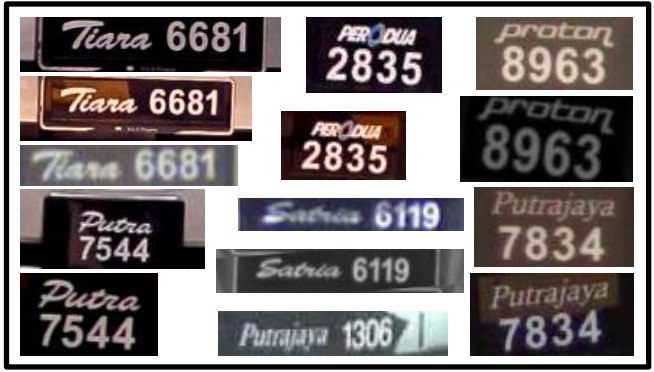

Figure 6: Samples images for algorithm testing

The proposed method will be assessed through three methods, which are true positive rate (TPR), false positive rate (FPR) and false negative rate (FNR). TPR means the percentage of special plates which are correctly identified and recognized. FPR means the percentage of normal plates are identified as special plates or special plates are wrongly recognized. FNR means the percentage of special plates are wrongly identified and recognized.

Out of 150 special plate images, a total number of 122 images are detected and recognized correctly as special plate. For the remaining 28 images, they are wrongly classified as normal license plate. While for the 350 standard license plate, none of them is detected and recognized as special plate group.

Table 1 shows the summarized result of the proposed method.

|  | TPR | FPR | FNR |
|---|---|---|---|
| Special Plates | 81.33% | 0% | 18.67% |

Table 1: summarized result of the proposed method

There is 18.67% of false negative result. This is due to condition of the detected plate image are too blurred or too small where SIFT features are difficult to be extracted for matching process. Figure 7 shows some sample images that are wrongly identified.

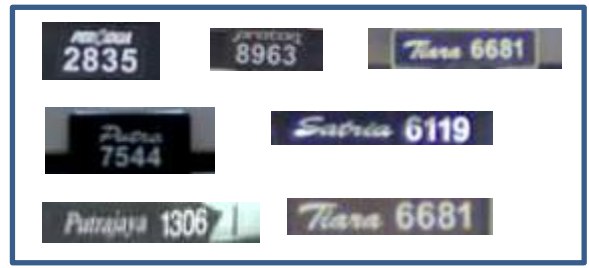

Figure 7: Samples images failed being identified as special plates

## V. Conclusion

This paper proposed a method to detect and recognize special license plate in Malaysia based on SIFT feature points. Experimental results showed that the proposed algorithm remains relatively robust in detecting and recognizing Malaysian special license plate which consists of cursive, connected, and italic letters. Although the proposed algorithm achieved 81.33% of TPR, blurring and low resolution images cannot be dealt successfully. Our next goal is to do further enhancement on robustness and stability of SIFT feature points under various complex conditions.